\documentclass{article}

\usepackage{PRIMEarxiv}

\usepackage[utf8]{inputenc} 
\usepackage[T1]{fontenc}    
\usepackage{hyperref}       
\usepackage{url}            
\usepackage{booktabs}       
\usepackage{amsfonts}       
\usepackage{nicefrac}       
\usepackage{microtype}      
\usepackage{lipsum}
\usepackage{fancyhdr}       
\usepackage{graphicx}       
\usepackage{float}
\graphicspath{{media/}}     

\title{SSSegmenation: An Open Source Supervised Semantic Segmentation Toolbox Based on PyTorch}

\author{
  Zhenchao Jin \\
  The University of Hong Kong \\
  \texttt{blwx96@connect.hku.hk} 
}

\begin{document}
\maketitle

\begin{abstract}
  This paper presents SSSegmenation, which is an open source supervised semantic image segmentation toolbox based on PyTorch.
  The design of this toolbox is motivated by MMSegmentation while it is easier to use because of fewer dependencies and achieves superior segmentation performance under a comparable training and testing setup.
  Moreover, the toolbox also provides plenty of trained weights for popular and contemporary semantic segmentation methods, including Deeplab, PSPNet, OCRNet, MaskFormer, \emph{etc}.
  We expect that this toolbox can contribute to the future development of semantic segmentation.
  Codes and model zoos are available at \href{https://github.com/SegmentationBLWX/sssegmentation/}{SSSegmenation}.
\end{abstract}


\section{Introduction}

Semantic image segmentation is a task of grouping pixels with different semantic categories.
Prior to SSSegmenation, many efforts have been made to provide high-quality codebases and unified benchmarks for semantic segmentation, \emph{e.g.}, MMSegmentation \cite{mmseg2020}, OpenSeg \cite{openseg2019} and PyTorch-Encoding \cite{pytorchencoding2017}.
Despite this, most of the codebases have not been updated for a long time and thus, lack support for the state-of-the-art segmentation algorithms.
Furthermore, the actively updated MMSegmentation toolbox contains complex mmlab-series dependencies, which is sometimes unfriendly to the researchers.
To reduce the learning cost of using a segmentation toolbox and provide high-quality re-implementation of the state-of-the-art segmentation approaches, we build SSSegmenation, an open source supervised semantic image segmentation toolbox based on PyTorch \cite{paszke2017automatic}.
The major features of SSSegmenation can be summarized as follows,
\begin{itemize}
  \item \textbf{High Performance.} The performance of re-implemented semantic segmentation algorithms is better than or comparable to other codebases.
  \item \textbf{Unified Benchmark.} Like conventional MMSegmentation \cite{mmseg2020}, various segmentation methods are unified into several specific modules. Benefiting from this design, SSSegmentation can integrate a great deal of popular and contemporary semantic segmentation frameworks and then, train and test them on unified benchmarks.
  \item \textbf{Fewer Dependencies.} SSSegmenation tries its best to avoid introducing more dependencies when reproducing novel semantic segmentation approaches.
\end{itemize}
SSSegmenation project is under active development and, we expect that this toolbox can facilitate the future development of semantic image segmentation.

\section{Supported Frameworks}

SSSegmenation supports abounding backbone networks (\emph{i.e.}, encoder) and segmentors (\emph{i.e.}, decoder).
The integrated backbone networks and segmentors in SSSegmenation are listed in this section.
Note that, most of the re-implementation algorithms in SSSegmenation are modified from the released codes in official repositories or MMSegmentation \cite{mmseg2020}. 

\subsection{Supported Backbone Networks}

\begin{itemize}
  \item \textbf{UNet \cite{ronneberger2015u}.} A classic backbone network in medical image segmentation, which is published in MICCAI 2015.
  \item \textbf{BEiT \cite{bao2021beit}.} A self-supervised bidirectional encoder vision representation model inspired by BERT \cite{devlin2018bert}, and it is accepted by ICLR 2022.
  \item \textbf{Twins \cite{chu2021twins}.} A variant of vision transformer architectures for visual tasks, which appears in NeurIPS 2021.
  \item \textbf{CGNet \cite{wu2020cgnet}.} A lightweight and efficient network for semantic segmentation, which is accepted by TIP 2020.
  \item \textbf{HRNet \cite{wang2020deep}.} A high-resolution representation learning network, and it appears in TPAMI 2020.
  \item \textbf{ERFNet \cite{romera2017erfnet}.} An efficient residual factorized convolution network for real-time semantic segmentation, which is published in T-ITS 2017.
  \item \textbf{ResNet \cite{he2016deep}.} A classic backbone network in image classification, which is accepted by CVPR 2016.
  \item \textbf{ResNeSt \cite{zhang2022resnest}.} A split-attention network for visual tasks, which appears in CVPRW 2022.
  \item \textbf{ConvNeXt \cite{liu2022convnet}.} A ViT-inspired convolution network, and it is published in CVPR 2022.
  \item \textbf{FastSCNN \cite{poudel2019fast}.} A fast segmentation convolutional neural network reported in	ArXiv 2019.
  \item \textbf{BiSeNetV1 \cite{yu2018bisenet}.} A real-time bilateral segmentation network, which is accepted by ECCV 2018.
  \item \textbf{BiSeNetV2 \cite{yu2021bisenet}.} A bilateral network with guided aggregation for real-time semantic segmentation, which is published in IJCV 2021.
  \item \textbf{MobileNetV2 \cite{sandler2018mobilenetv2}.} A widely used lightweight and real-time convolution network for various visual applications, which is accepted by CVPR 2018.
  \item \textbf{MobileNetV3 \cite{howard2019searching}.} A lightweight and real-time convolution network based on a combination of complementary search techniques as well as a novel architecture design, which appears in ICCV 2019.
  \item \textbf{SwinTransformer \cite{liu2021swin}.} A widely used hierarchical vision transformer architecture using shifted windows for visual tasks, which is published in ICCV 2021.
  \item \textbf{VisionTransformer \cite{dosovitskiy2020image}.} The first vision transformer architecture, which is accepted by IClR 2021.
\end{itemize}

\subsection{Supported Segmentors}

\begin{itemize}
  \item \textbf{FCN \cite{long2015fully}.} A conventional encoder-decoder segmentation framework, and it is published in TPAMI 2017.
  \item \textbf{CE2P \cite{ruan2019devil}.} A classic human parsing framework, which appears in AAAI 2019.
  \item \textbf{SETR \cite{zheng2021rethinking}.} A pure transformer structure (\emph{i.e.}, without convolution and resolution reduction) to encode an image as a sequence of patches, which is proposed in CVPR 2021.
  \item \textbf{ISNet \cite{jin2021isnet}.} A high-performance semantic segmentation framework utilizing image-level and semantic-level contextual information to augment pixel representations, which is proposed in ICCV 2021.
  \item \textbf{ICNet \cite{zhao2018icnet}.} An image cascade network that incorporates multi-resolution branches under proper label guidance to reduce a large portion of computation for pixel-wise label inference, which is introduced in ECCV 2018.
  \item \textbf{CCNet \cite{huang2019ccnet}.} A criss-cross network for effectively and efficiently obtaining full-image contextual information, which is accepted by ICCV 2019.
  \item \textbf{DANet \cite{fu2019dual}.} A dual attention network to adaptively integrate local features with their global dependencies, which is published in	CVPR 2019.
  \item \textbf{DMNet \cite{he2019dynamic}.} A dynamic multi-scale neural network to adaptively aggregate multi-scale contents for predicting pixel-level semantic labels, which is proposed in ICCV 2019.
  \item \textbf{GCNet \cite{cao2019gcnet}.} A segmentation framework that unifies non-local block \cite{wang2018non} and squeeze-excitation network \cite{hu2018squeeze}  into a three-step general framework for global context modeling, which is proposed in TPAMI 2020.
  \item \textbf{ISANet \cite{huang2019interlaced}.} An interlaced sparse self-attention approach to improve the efficiency of self-attention mechanism for semantic segmentation, which is accepted by IJCV 2021.
  \item \textbf{EncNet \cite{zhang2018context}.} A context encoding module to explore the impact of global contextual information in semantic segmentation, which is presented in CVPR 2018.
  \item \textbf{OCRNet \cite{yuan2020object}.} A semantic segmentation framework that emphasizes utilizing object-contextual representations to enhance pixel representations, which is proposed in ECCV 2020.
  \item \textbf{DNLNet \cite{yin2020disentangled}.} A disentangled non-local block to facilitate learning for both the whitened pairwise term and the unary term in semantic segmentation, which appears in ECCV 2020.
  \item \textbf{ANNNet \cite{zhu2019asymmetric}.} An asymmetric non-local neural network (consisting of asymmetric pyramid non-local block and asymmetric fusion non-local block) for semantic segmentation, which is accepted by ICCV 2019.
  \item \textbf{EMANet \cite{li2019expectation}.} An expectation-maximization attention module for aggregating contextual information in a robust way, which is proposed in ICCV 2019.
  \item \textbf{PSPNet \cite{zhao2017pyramid}.} A pyramid pooling module is designed to capture multi-scale contextual information, which is accepted by CVPR 2017.
  \item \textbf{PSANet \cite{zhao2018psanet}.} A point-wise spatial attention network to relax the local neighborhood constraint for semantic segmentation, which is introduced in CVPR 2017.
  \item \textbf{APCNet \cite{he2019adaptive}.} An adaptive pyramid context network for semantic segmentation, accepted by CVPR 2019.
  \item \textbf{FastFCN \cite{wu2019fastfcn}.} A joint pyramid upsampling module to formulate the task of extracting high-resolution feature maps into a joint upsampling problem, which is reported in	ArXiv 2019.
  \item \textbf{UPerNet \cite{xiao2018unified}.} A hierarchical structure to learn from heterogeneous data from multiple image datasets in semantic segmentation, which is proposed in ECCV 2018.
  \item \textbf{PointRend \cite{kirillov2020pointrend}.}  Taking image segmentation as a rendering problem, proposed in CVPR 2020.
  \item \textbf{Deeplabv3 \cite{chen2017rethinking}.} A segmentation framework that uses atrous convolution in cascade or in parallel to capture multi-scale context by adopting multiple atrous rates, which is introduced in CVPR 2017.
  \item \textbf{Segformer \cite{xie2021segformer}.} A simple, efficient yet powerful semantic segmentation framework which unifies transformers with lightweight multilayer perceptron (MLP) decoders, proposed in NeurIPS 2021.
  \item \textbf{MaskFormer \cite{cheng2021per}.} A simple mask classification model that formulates semantic segmentation as a per-pixel classification task, which is published in NeurIPS 2021.
  \item \textbf{SemanticFPN \cite{kirillov2019panoptic}.} A panoptic feature pyramid network applied in semantic segmentation, which is proposed in CVPR 2019.
  \item \textbf{NonLocalNet \cite{wang2018non}.} A non-local block to capture long-range dependencies for semantic segmentation, which is accepted by CVPR 2018.
  \item \textbf{Deeplabv3Plus \cite{chen2018encoder}.} An extended version of Deeplabv3, which is published in ECCV 2018.
  \item \textbf{MCIBI \cite{jin2021mining}.} A semantic segmentation framework targeting mining contextual information beyond image, which is accepted by ICCV 2021.
  \item \textbf{MCIBI++ \cite{jin2022mcibi++}.} An extended version of MCIBI, which is published in TPAMI 2022.
  \item \textbf{Mixed Precision (FP16) Training \cite{micikevicius2017mixed}.} Mixed precision training aims at training deep neural networks using half precision floating point (FP16) numbers, and it is proposed in ArXiv 2017.
\end{itemize}

\section{Benchmarks}

The segmentation frameworks integrated into SSSegmenation can be trained on various popular benchmark datasets, including LIP \cite{gong2017look}, ADE20k \cite{zhou2017scene}, CityScapes \cite{cordts2016cityscapes}, COCO Stuff \cite{caesar2018coco}, Pascal Context \cite{mottaghi2014role}, VSPW \cite{miao2021vspw}, to name a few.
Table \ref{tab:reportedperformance} shows the performance of various segmentation methods on Pascal VOC, ADE20K and Cityscapes datasets.
The readers can refer to the official repository of SSSegmenation for training details and more segmentation results.

\begin{table}[t]
\centering
\caption{
  The performance of different segmentation approaches on three popular benchmark datasets.
  All results are obtained under a single-scale setting without extra tricks, \emph{e.g.}, flipping and multi-scale testing.
}\label{tab:reportedperformance}
\resizebox{1.00\textwidth}{!}{
    \begin{tabular}{c|c|c|c|c|c}
    \toprule
    Method                               &Backbone                     &Stride        &ADE20K (\emph{train / val})           &Pascal VOC (\emph{train / test})     &Cityscapes (\emph{train / val})      \\
    \hline       
    ANNNet                               &ResNet-50                    &$8\times$     &41.75                                 &76.68                                &78.36                                \\
    ANNNet                               &ResNet-101                   &$8\times$     &43.98                                 &78.15                                &79.34                                \\
    \hline
    APCNet                               &ResNet-50                    &$8\times$     &43.47                                 &76.97                                &79.02                                \\
    APCNet                               &ResNet-101                   &$8\times$     &45.74                                 &78.99                                &79.71                                \\
    \hline
    CCNet                                &ResNet-50                    &$8\times$     &42.47                                 &77.43                                &79.15                                \\
    CCNet                                &ResNet-101                   &$8\times$     &44.00                                 &78.02                                &80.08                                \\
    \hline
    DANet                                &ResNet-50                    &$8\times$     &42.90                                 &76.39                                &79.47                                \\
    DANet                                &ResNet-101                   &$8\times$     &44.37                                 &77.97                                &80.55                                \\
    \hline
    Deeplabv3                            &ResNet-50                    &$8\times$     &43.19                                 &77.72                                &79.62                                \\
    Deeplabv3                            &ResNet-101                   &$8\times$     &45.16                                 &79.52                                &80.28                                \\
    \hline
    Deeplabv3Plus                        &ResNet-50                    &$8\times$     &44.51                                 &77.43                                &80.38                                \\
    Deeplabv3Plus                        &ResNet-101                   &$8\times$     &45.72                                 &79.19                                &81.09                                \\
    Deeplabv3Plus                        &ResNeSt-101                  &$8\times$     &46.48                                 &79.76                                &80.30                                \\
    \hline
    DMNet                                &ResNet-50                    &$8\times$     &43.54                                 &77.38                                &79.17                                \\
    DMNet                                &ResNet-101                   &$8\times$     &45.53                                 &79.15                                &79.90                                \\
    \hline
    DNLNet                               &ResNet-50                    &$8\times$     &43.50                                 &76.73                                &79.75                                \\
    DNLNet                               &ResNet-101                   &$8\times$     &44.88                                 &78.37                                &80.64                                \\
    \hline
    EMANet                               &ResNet-50                    &$8\times$     &41.77                                 &75.29                                &77.96                                \\
    EMANet                               &ResNet-101                   &$8\times$     &44.39                                 &76.43                                &79.54                                \\
    \hline
    EncNet                               &ResNet-50                    &$8\times$     &40.60                                 &75.53                                &77.98                                \\
    EncNet                               &ResNet-101                   &$8\times$     &43.43                                 &77.61                                &78.70                                \\
    \hline
    GCNet                                &ResNet-50                    &$8\times$     &42.53                                 &76.50                                &78.69                                \\
    GCNet                                &ResNet-101                   &$8\times$     &44.19                                 &78.81                                &79.93                                \\
    \hline
    ICNet                                &ResNet-50                    &$8\times$     &-                                     &-                                    &76.60                                \\
    ICNet                                &ResNet-101                   &$8\times$     &-                                     &-                                    &76.27                                \\
    \hline
    ISANet                               &ResNet-50                    &$8\times$     &42.60                                 &76.99                                &79.34                                \\
    ISANet                               &ResNet-101                   &$8\times$     &44.08                                 &78.60                                &80.58                                \\
    \hline
    ISNet                                &ResNet-50                    &$8\times$     &44.22                                 &-                                    &79.32                                \\
    ISNet                                &ResNet-101                   &$8\times$     &45.92                                 &-                                    &80.56                                \\
    \hline
    Deeplabv3+MCIBI                      &ResNet-50                    &$8\times$     &44.39                                 &-                                    &-                                    \\
    Deeplabv3+MCIBI                      &ResNet-101                   &$8\times$     &45.66                                 &-                                    &-                                    \\
    \hline
    UPerNet+MCIBI++                      &ResNet-50                    &$8\times$     &44.30                                 &79.48	                               &-                                    \\
    UPerNet+MCIBI++                      &ResNet-101                   &$8\times$     &46.39                                 &80.42                                &-                                    \\
    \hline
    NonLocalNet                          &ResNet-50                    &$8\times$     &42.15                                 &77.08                                &78.34                                \\
    NonLocalNet                          &ResNet-101                   &$8\times$     &44.49                                 &78.89                                &80.42                                \\
    \hline
    OCRNet                               &ResNet-50                    &$8\times$     &42.47                                 &76.75                                &79.40                                \\
    OCRNet                               &ResNet-101                   &$8\times$     &43.99                                 &78.82                                &80.61                                \\
    OCRNet                               &HRNetV2p-W48                 &$4\times$     &44.03                                 &77.60                                &81.44                                \\
    \hline
    PointRend                            &ResNet-50                    &$8\times$     &37.80                                 &69.84                                &76.89                                \\
    PointRend                            &ResNet-101                   &$8\times$     &40.26                                 &72.31                                &78.80                                \\
    \hline
    PSANet                               &ResNet-50                    &$8\times$     &41.99                                 &77.06                                &78.88                                \\
    PSANet                               &ResNet-101                   &$8\times$     &43.85                                 &78.97                                &79.65                                \\
    \hline
    PSPNet                               &ResNet-50                    &$8\times$     &42.64                                 &77.93                                &79.05                                \\
    PSPNet                               &ResNet-101                   &$8\times$     &44.55                                 &79.04                                &79.94                                \\
    \hline
    FCN                                  &ResNet-50                    &$8\times$     &36.96                                 &67.80                                &75.16                                \\
    FCN                                  &ResNet-101                   &$8\times$     &41.22                                 &70.59                                &76.31                                \\
    FCN                                  &BiSeNetV1+ResNet-18          &$32\times$    &-                                     &-                                    &75.76                                \\
    FCN                                  &BiSeNetV1+ResNet-50          &$32\times$    &-                                     &-                                    &77.78                                \\
    FCN                                  &BiSeNetV2                    &$32\times$    &-                                     &-                                    &74.62                                \\
    FCN                                  &CGNet-M3N21                  &$8\times$     &-                                     &-                                    &68.53                                \\
    FCN                                  &ERFNet                       &$8\times$     &-                                     &-                                    &76.44                                \\
    \hline
    UPerNet                              &ResNet-50                    &$8\times$     &43.02                                 &76.86                                &79.08                                \\
    UPerNet                              &ResNet-101                   &$8\times$     &44.92                                 &79.13                                &80.39                                \\
    UPerNet                              &BEiT-B                       &$16\times$    &53.22                                 &-                                    &-                                    \\
    UPerNet                              &BEiT-L                       &$16\times$    &56.52                                 &-                                    &-                                    \\
    UPerNet                              &ConvNeXt-L-21k               &$8\times$     &53.41                                 &-                                    &-                                    \\
    UPerNet                              &ConvNeXt-XL-21k              &$8\times$     &53.68                                 &-                                    &-                                    \\
    UPerNet                              &Swin-Small                   &$32\times$    &48.39                                 &-                                    &-                                    \\
    UPerNet                              &Swin-Base                    &$32\times$    &51.02                                 &-                                    &-                                    \\
    UPerNet                              &SVT-Large                    &$32\times$    &49.80                                 &-                                    &-                                    \\
    UPerNet                              &PCPVT-Large                  &$32\times$    &49.35                                 &-                                    &-                                    \\
    \hline
    MaskFormer                           &Swin-Small                   &$32\times$    &49.91                                 &-                                    &-                                    \\
    MaskFormer                           &Swin-Base                    &$32\times$    &53.22                                 &-                                    &-                                    \\
    \hline
    SETR-Naive                           &ViT-Large                    &$32\times$    &48.43                                 &-                                    &-                                    \\
    SETR-PUP                             &ViT-Large                    &$32\times$    &48.51                                 &-                                    &-                                    \\
    SETR-MLA                             &ViT-Large                    &$32\times$    &49.61                                 &-                                    &-                                    \\
    \bottomrule
\end{tabular}}
\vspace{-0.4cm}
\end{table}

\section{Code Structure of SSSegmenation}

SSSegmenation mainly consists of \emph{dataset module}, \emph{parallel module}, \emph{backbone module}, \emph{segmentor module}, \emph{loss module}, \emph{optimizer module} and \emph{scheduler module}.
Specifically, \emph{dataset module} is defined to load training and testing images.
\emph{parallel module} simply call the $nn.parallel.DistributedDataParallel$ function in PyTorch to achieve distributed training.
\emph{backbone module} is employed to define the encoder structure while \emph{segmentor module} is leveraged to define the decoder structure.
\emph{loss module} is introduced to define the objective functions for semantic segmentation.
\emph{optimizer module} and \emph{scheduler module} are designed for optimizing the network parameters through back propagation algorithm.
Furthermore, it is also convenient to integrate more segmentation algorithms into SSSegmenation since we only need to define our algorithm in the corresponding module folder and import it in the $builder.py$ in the folder.

The minimum requirement for utilizing SSSegmenation is a CUDA development environment with PyTorch and some necessary but easy-to-installed packages, including \emph{chainercv}, \emph{pillow}, \emph{opencv-python}, \emph{pandas}, \emph{cython}, \emph{numpy}, \emph{scipy}, \emph{tqdm} and \emph{argparse}.
Also, you can install \emph{timm}, \emph{mmcv} and other packages to extend the functions of SSSegmenation.

In summary, SSSegmenation is a simple yet effective codebase for semantic image segmentation.
We hope this toolbox can further advance researches on semantic image segmentation.

\bibliographystyle{unsrt}  
\bibliography{references}

\end{document}